\documentclass[runningheads]{llncs}
\usepackage[T1]{fontenc}
\usepackage{graphicx}
\usepackage[table]{xcolor}
\usepackage{textcomp}
\usepackage{amsmath,amssymb,amsfonts}
\usepackage{bm}
\usepackage{cite}
\usepackage{booktabs}
\usepackage{multirow}
\usepackage{makecell}
\usepackage{subfig}
\usepackage{orcidlink} 
\usepackage{afterpage}
\usepackage{algpseudocode}
\usepackage{algorithm}
\usepackage{rotating}
\usepackage{nicematrix}
\usepackage{amssymb}
\usepackage{capt-of}
\usepackage{mathabx}
\usepackage{enumitem}

\newcommand*{\twoelementtable}[3][l]%
{%
    \begin{tabular}[t]{@{}#1@{}}%
        #2\tabularnewline
        #3%
    \end{tabular}%
}
\newcolumntype{x}{>{\centering\arraybackslash}p{.75cm}}
\newcolumntype{C}[1]{>{\centering\arraybackslash}p{#1}}
\makeatletter
\g@addto@macro{\endtabular}{\rowfont{}}
\makeatother
\newcommand{\rowfonttype}{}
\newcommand{\rowfont}[1]{
   \gdef\rowfonttype{#1}#1%
}
\usepackage{pifont}%
\newcommand{\cmark}{\textcolor{teal}{\ding{51}}}%
\newcommand{\xmark}{\textcolor{red}{\ding{55}}}%
 \newcommand{\textapproxx}{{\raise.17ex\hbox{$\scriptstyle\mathtt{\sim}$}}}

\usepackage{xspace}
\makeatletter
\DeclareRobustCommand\onedot{\futurelet\@let@token\@onedot}
\def\@onedot{\ifx\@let@token.\else.\null\fi\xspace}

\def\eg{\emph{e.g}\onedot} 
\def\ie{\emph{i.e}\onedot} 
\def\cf{\emph{c.f}\onedot} 
 
\def\wrt{w.r.t\onedot}

\makeatletter
\def\mathcolor#1#{\@mathcolor{#1}}
\def\@mathcolor#1#2#3{%
  \protect\leavevmode
  \begingroup\color#1{#2}#3\endgroup
}
\makeatother

\usepackage{titlesec}
\titlespacing*{\section} {00pt}{2.5ex}{1.5ex}
\titlespacing*{\subsection} {0pt}{2ex}{1ex}
\begin{document}
\title{Few-Shot 3D Volumetric Segmentation \\with Multi-Surrogate Fusion}

\author{Meng Zheng\thanks{Corresponding author.}\inst{1}\orcidlink{0000-0002-6677-2017} \and
Benjamin Planche\inst{1}\orcidlink{0000-0002-6110-6437} \and
Zhongpai Gao\inst{1}\orcidlink{0000-0003-4344-4501} \and \\
Terrence Chen\inst{1} \and
Richard J. Radke\inst{2}\orcidlink{0000-0001-5064-7775} \and
Ziyan Wu\inst{1}\orcidlink{0000-0002-9774-7770}
}

\authorrunning{M. Zheng et al.}
%
\institute{United Imaging Intelligence, Boston, MA, USA\\
\email{\{first.last\}@uii-ai.com}\\
\and
Rensselaer Polytechnic Institute, Troy, NY, USA\\
\email{rjradke@ecse.rpi.edu}
}

\maketitle              

\begin{abstract}
Conventional 3D medical image segmentation methods typically require learning heavy 3D networks (\eg, 3D-UNet), as well as large amounts of in-domain data with accurate pixel/voxel-level labels to avoid overfitting. These solutions are thus extremely time- and labor-expensive, but also may easily fail to generalize to unseen objects during training.
To alleviate this issue, we present MSFSeg, a novel few-shot 3D segmentation framework with a lightweight multi-surrogate fusion (MSF). MSFSeg is able to automatically segment unseen 3D objects/organs (during training) provided with one or a few annotated 2D slices or 3D sequence segments, via learning dense query-support organ/lesion anatomy correlations across patient populations.
Our proposed MSF module mines comprehensive and diversified morphology correlations between unlabeled and the few labeled slices/sequences through multiple designated surrogates, making it able to generate accurate cross-domain 3D segmentation masks given annotated slices or sequences. 
We demonstrate the effectiveness of our proposed framework by showing superior performance on conventional few-shot segmentation benchmarks compared to prior art, and remarkable cross-domain cross-volume segmentation performance on proprietary 3D segmentation datasets for challenging entities, \ie, tubular structures, with only limited 2D or 3D labels.

\end{abstract}

\section{Introduction}
Volumetric/3D medical image segmentation is crucial to various clinical applications, from treatment planning to disease monitoring \cite{hatamizadeh2022unetr,wang2019volumetric,xie2021cotr}, and has been benefiting from  advances in data-centric machine-learning solutions. 
However, densily annotating the large-scale volumetric data required for their fully-supervised training is extremely time/labor-consuming, and the generalizability of these models to unseen object classes is uncertain. 

\begin{figure*}[t]
    \centering
    \includegraphics[width=1\linewidth]{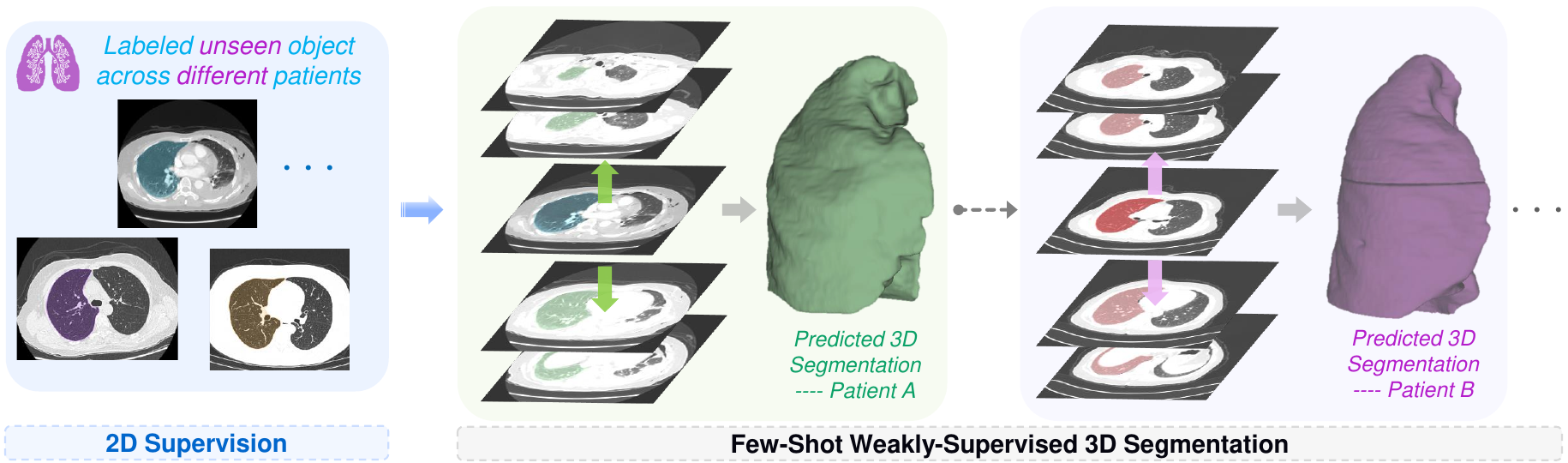}
    \caption{Proposed few-shot weakly-supervised 3D segmentation pipeline.}
    \label{fig:teaser}
\end{figure*}

To tackle these limitations, the task of Few Shot Segmentation (FSS) has gained traction \cite{shaban2017one,li2021adaptive,liu2020crnet,ouyang2020self,roy2020squeeze,wang2019panet,CRAPNet_WACV23,RPT_MICCAI23,VQV_CVPR23,RAP_MIA23,ODEs_TMI23,CAT_MICCAI23}. FSS aims to develop domain-generalizable neural networks capable of segmenting previously unseen objects from new \textit{query} images/slices with minimal annotated examples (\ie, \textit{support} images/slices and masks), improving data annotation efficiency.

Efforts to tackle FSS tasks have focused on both class-wise \cite{aaai2019_attention,Zhang2019CANetCS,zhang2020sg} and pixel-wise \cite{min2021hypercorrelation,lang2022learning,shi2022DCAMA,PGN_ICCV19,DAN_ECCV20} prototyping architectures that rely on computing similarity/affinity correlation between support and query data. However, few-shot 3D segmentation, which involves segmenting video or volumetric data with limited labeled 2D/3D images \cite{roy2020squeeze,scribble_TMI23,Kim_An_Chikontwe_Park_2021}, remains underexplored as proposed networks need fully-labeled 3D supervisions, compromising generalizability and computation efficiency. 
%
Existing few-shot 3D-segmentation solutions \cite{roy2020squeeze,ouyang2020self,CRAPNet_WACV23,SRCL_MICCAI22,CAT_ICIT22} often rely on 2D FSS networks that treat each slice individually, neglecting patient-specific inter-slice morphology correlations crucial for accurate 3D mask prediction (\cf useful clues and strong constraints).  
%
Moreover, they typically only consider \textit{1-shot} cases~\cite{ouyang2020self,CRAPNet_WACV23,SRCL_MICCAI22,CAT_ICIT22,VQV_CVPR23} and overlook multi-support ones ($n$\textit{-shot}), or use generic aggregation methods (\eg, ensemble \cite{self_guide_CVPR} or voting strategies \cite{RAP_MIA23,min2021hypercorrelation,shaban2017one}), which may compromise diverse morphology correlations across support image/mask or sequence/mask pairs. 
%
Effective aggregation of information from multiple support images or sequences is crucial for few-shot 3D segmentation, as 3D volumes contain spatially correlated 2D slices, and different patient volumes exhibit distinct morphology correlations. Utilizing both consistent and diversified inter-slice/mask morphology variations across patients, given limited labeled support, is key to solving this problem.

We thus introduce \textbf{MSFSeg}, a novel few-shot 3D segmentation pipeline with a Multi-Surrogate Fusion (MSF)-informed few-shot network (\cf Fig.\ \ref{fig:teaser}). MSFSeg can segment unseen 3D objects with just a few support 2D slices or sequences. It accommodates various support formats and numbers ($n$-shot, $n\geq1$), \ie, taking 2D slices or sequential slice segments from different scans as support. Exploring dense foreground/background pixel relationships and mining semantic features across multiple supports from diverse patient data distributions, MSFSeg enhances query mask prediction. 
%
We evaluate MSFSeg on conventional abdominal CT and MRI organ segmentation benchmarks, comparing it to prior FSS methods. Additionally, we demonstrate its effectiveness on few-shot weakly-supervised 3D segmentation across 4 organs and 40 data volumes collected from a hospital.
To summarize, our key contributions include:
\begin{itemize}[topsep=0pt,wide=2pt]
    \item We introduce a new few-shot 3D segmentation pipeline for unseen 3D objects, given limited labeled 2D or 3D supports as weak supervision. It  leverages intra- and inter-patient/volume morphology correlations to better predict 3D masks.
    
    \item We design a novel FSS network with multi-surrogate fusion to exhaustively explore and mine complementary/diversified semantic information across various support slices/sequences, resulting in superior accuracy in transferred domains.

    \item We demonstrate the effectiveness and generalizability of MSFSeg, sharing state-of-the-art FSS performance on medical benchmarks, as well as satisfying few-shot 3D segmentation performance on clinical data, particularly in segmenting challenging objects like tubular structures in cross-domain settings.
\end{itemize}

\section{Methodology}
\subsection{MSFSeg with Multi-Surrogate Fusion} 
\label{sec:Architecture}
Considering the task of segmenting unseen objects in 3D medical images given a few support slices or sequences (akin to transfer learning), we base the proposed MSFSeg on existing few-shot segmentation pipelines \cite{min2021hypercorrelation,roy2020squeeze}, which takes one 2D query image $\mathbf{I}^{\text{q}} \in \mathbb{R}^{h \times w}$ of size $h \times w$, $n$ support sequences $\{\mathbf{I}^{\text{s}}_i \in \mathbb{R}^{d_i \times h \times w}\}_{i=1}^n$ ($d_i$ is the number of slices in each sequence, $d_i=1$ is same as conventional 2D FSS where supports are single slices),
along with their corresponding labeled segmentation masks
$\{\mathbf{M}^{\text{s}}_i \in \mathbb{R}^{d_i \times h \times w}\}_{i=1}^n$
to infer the mask $\mathbf{M}^{\text{q}}$ of the query image. 
Unlike prior art that considers only one support (1-shot) \cite{RPNet_ICCV21,SRCL_MICCAI22,ouyang2020self,location_ISBI21,CRAPNet_WACV23,roy2020squeeze} or applies vanilla aggregation strategies for $n$-shot cases at inference time only \cite{min2021hypercorrelation,shaban2017one}, our solution takes $n$ support slices at training time for end-to-end optimization, leveraging the proposed multi-surrogate fusion.
Inherently optimized for leveraging diverse multi-support information, our proposed MSFSeg can effectively segment unseen 3D objects across different patients (query volumes).


\noindent\textbf{A) Feature Extraction and Self-Attention.} 
We follow existing 1-shot FSS pipelines \cite{RPNet_ICCV21,SRCL_MICCAI22,ouyang2020self,location_ISBI21,CRAPNet_WACV23,roy2020squeeze,hong2022cost} and apply a pretrained ResNet-101 \cite{resnet_CVPR16} 
to extract query and support image features. We collect the sets of multi-scale query and support feature maps, 
$\{\mathbf{f}^{\text{q}}_j\}_{j=1}^b$ 
and $\{\mathbf{f}^{\text{s}}_{1..N,j}\}_{j=1}^b$ ($N=\sum^n_{i=1}d_i$ is the total number of slices from $n$ supports), out of the $b$ ResNet blocks, from query image $\mathbf{I}^{\text{q}}$ and support sequences $\mathbf{I}^{\text{s}}_{1..n}$ respectively.
At each feature scale level, we then compute the multi-head attention  (\cf Transformer strategy \cite{vaswani2017attention}) to correlate query and support pixel-wise morphology and predict $n$ coarse query mask features $\{\mathbf{\hat{M}}^{\text{q}}_{1..n,j}\}_{j=1}^b$ (at $b$ different scales) given query image and support image/mask pairs. 
Specifically, each support mask $\mathbf{M}^{\text{s}}_i$, used as attention value vector, is first downsampled to the same size $h_j \times w_j$ as the considered feature map (\cf scaling by $j^{\text{th}}$ ResNet block resulting in $h_j \times w_j \times c_j$ features) and then flattened into a 1-dimensional vector to form $V_{i,j}$. We similarly resample the feature maps $\mathbf{f}^{\text{q}}_j$ and $\mathbf{f}^{\text{s}}_{1..n,j}$ into $(h_j*w_j) \times c_j$ matrices, with each pixel treated as a token. Adding positional encoding and linear projections, we respectively get the attention query $Q_j$ and key $K_{i,j}$ matrices. 
Note that $d_i$ support feature maps and masks from the same support sequence $\mathbf{I}^{\text{s}}_{i}$ are concatenated and treated as different tokens to form the key and value matrices. This way, the following multi-head attention mechanism can internally aggregate token-wise information across $d_i$ slices via scaled dot-product attention and enable flexible support sequence length.

The multi-head attention is finally computed for each support pair $i$ and scale $j$ to generate $n \times b$ initial multi-scale query masks $\{\mathbf{\hat{M}}^{\text{q}}_{1..n,j}\}_{j=1}^b$: 
$
\small{\operatorname{Attention}(Q_j,}$ 
$\small{K_{i,j}, V_{i,j})=\operatorname{Softmax}\Big(Q_j K_{i,j}^\intercal / \sqrt{d}\Big) V_{i,j}},
$
with $d$ dimension of $Q_j$.
The intuition behind is that by multiplying $Q$ and $K^\intercal$, pixel-wise correlations between query and support images are densely captured, to enhance and weigh the support mask towards predicting an accurate query mask. The inner product of $Q$ and $K^\intercal$ acts like a projection 
to bridge the pixel-wise image correlations in support-query data with the mask correlations.
We finally aggregate the $b$ multi-scale results via upsampling and summing, followed by cascaded 2D convolutions (details in appendix). This produces $n$ multi-scale-informed query mask features $\mathbf{\hat{M}}^{\text{q}}_{1..n}$. 

\begin{figure*}[t]
    \centering    
    \includegraphics[width=\linewidth]{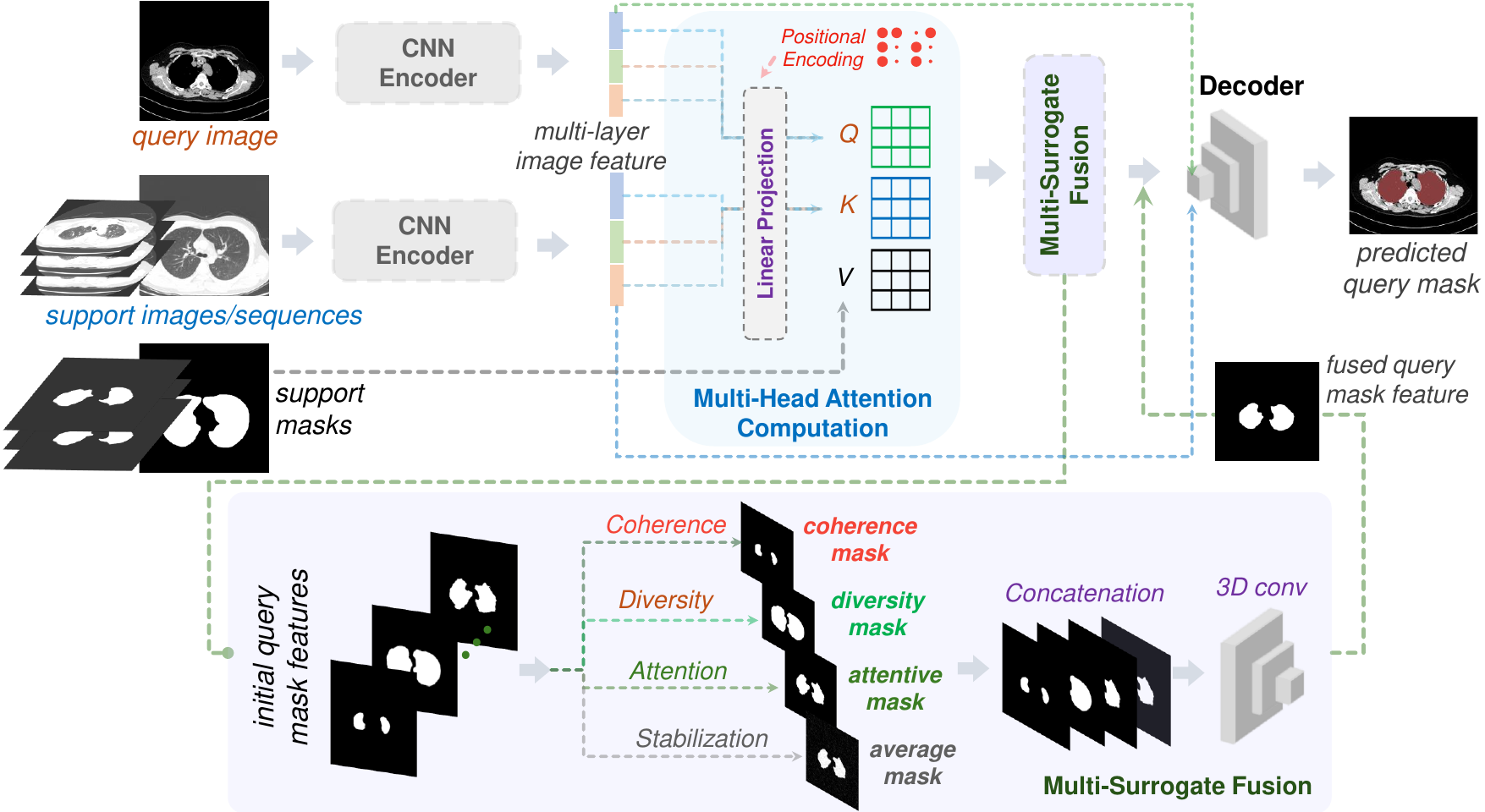}
    \caption{Proposed MSF solution for few-shot 3D segmentation.}
    \label{fig:fss_pipeline}
\end{figure*}

\noindent\textbf{B) Fusion via Mixture of Surrogates.} 
\label{sec:Multi-Surrogate}
Efficiently and effectively utilizing multi-support information has been a challenging, long-standing problem, especially 
for cross-domain cross-patient/volume segmentation in few-shot setting, where learning consistent and patient-specific morphology correlations across different patient data could greatly benefit FSS pipelines in generalizing better to unseen patient data and objects.
We thus propose a Multi-Surrogate Fusion (MSF) module (Fig.\ \ref{fig:fss_pipeline}, bottom) based on a ResNet-101 backbone, to effectively distill complementary morphology information across multiple supports.

Specifically, our MSF takes the $n$ multi-scale query mask features $\mathbf{\hat{M}}^{\text{q}}_{1..n}$, and assign 4 specialized surrogates to explore and undermine local and global morphology information that is (1) coherent, (2) diverse, (3) cross-attentive, and (4) stabilized across $n$ query mask features predicted from $n$ different support images/masks. Specifically, we deploy a coherence surrogate to distill consistent information  by performing soft intersection operation (element-wise product of soft-maxed masks) to generate coherence masks $\mathbf{\hat{M}}^{\text{q}}_{\text{coh}} = \bigcap^n_{i=1} \hat{\mathbf{M}}_i^q$, and extract diversified information by performing soft union operation over $n$ query mask features (sum of soft-maxed masks) to generate diversity masks, \ie $\mathbf{\hat{M}}^{\text{q}}_{\text{div}} = \bigcup^n_{i=1} \hat{\mathbf{M}}_i^q$. 
We then perform channel-wise attention across all query mask features, each query treated as one channel, aiming to find attentive masks: 
\begin{equation}
    \mathbf{M}^q_{ca} = \frac{1}{n}\sum_{i=1}^{n}\hat{\mathbf{M}}_i^q \otimes \operatorname{Softmax}\big(\operatorname{AvgPool3d}(\hat{\mathbf{M}}_i^q)\big),
\end{equation}
where ``$\otimes$" is the element-wise product. $\mathbf{M}^q_{ca}$ extracts and fuses complementary foreground information while suppressing distracting background information from multiple experts. 
We also average over $n$ query mask features, \ie $\mathbf{M}^q_{avg} = \frac{1}{n}\sum_{i=1}^{n}\hat{\mathbf{M}}_i^q$, to obtain stabilized global information across $\mathbf{\hat{M}}^{\text{q}}_{1..n}$.
Finally, the coherence, diversity, attentive, and average masks---resp. $\mathbf{\hat{M}}^{\text{q}}_{\text{coh}}$, $\mathbf{\hat{M}}^{\text{q}}_{\text{div}}$, $\mathbf{\hat{M}}^{\text{q}}_{\text{ca}}$, $\mathbf{\hat{M}}^{\text{q}}_{\text{avg}}$---are concatenated together channel-wise and input into a 3D conv.\ layer to fuse them into query mask features. While our coherence and diversity surrogates aim to explore complementary local, pixel-wise morphology query-support and support-support correlations, our attention and stability surrogates distill global information across supports for improved query mask prediction.

\noindent\textbf{C) Mask Upsampling and Refinement.} 
We finally concatenate the fused query mask to the image features $\mathbf{f}^{\text{q}}_1$ 
and $\mathbf{f}^{\text{s}}_{1..n,1}$ (skip-connection operations similar to U-Net \cite{unet_miccai}) and send them through two deconvolutional blocks (decoder) to refine and upsample into the final $h \times w$ query mask prediction. 

\subsection{Few-Shot 3D Segmentation Workflow}
Leveraging MSFSeg (Sec.\ \ref{sec:Architecture}), we define an automated few-shot 3D segmentation workflow, explained in Alg.\ \ref{algo:prop_workflow}. 
Specifically, given $B$ 3D volumes with an unseen object unlabeled, we first annotate one or a few slices/sequences (from one or different data volumes) as supports, then segment each of the $B$ data volumes by sending the query slice and labeled support slices/sequences to our MSFSeg for few-shot 3D segmentation. Note that we are able to constantly add segmented slices to the support pool with quality control filtering, to better segment the unlabeled slices/volumes by selecting reliable $n$ support slices (\ie, support slices with high-level features extracted from ResNet101 backbone which have highest cosine similarity to the query feature) out of the support pool.

\begin{algorithm}[t]
\caption{
MSFSeg-based few-shot 3D segmentation workflow.
}
\label{algo:prop_workflow}
\textbf{Input} Unseen object, unlabeled $B$ 3D volumes $\{V_k\}_{k=1}^{B}$, empty support pool $\mathcal{P}$.\\
\textbf{Output} $B$ labeled 3D volumes.
\begin{algorithmic}[1]
\State Label $n$ sequences $\mathbf{I}^{\text{s}}_{1..n}$ from $\{V_k\}_{k=1}^{B}$ as exemplar/support, and generate label $\mathbf{M}^{\text{s}}_{1..n}$.
\State Add labeled slices to support pool, $\mathcal{P} \gets (\mathbf{I}^{\text{s}}_{1..n}, \mathbf{M}^{\text{s}}_{1..n})$.
\For {${\text{volume } V_k \in V_{1..B}}$}
\For {${\text{2D slice } \mathbf{I}^{\text{q}}_{k,i} \in \text{volume } V_k}$}
\State Identify $n$ most similar support slices from pool: $(\mathbf{I}^{\text{s}}_{1..n}, \mathbf{M}^{\text{s}}_{1..n}) \in \mathcal{P}$.
\State Generate label for $\mathbf{I}^{\text{q}}_{k,i}$: $\mathbf{M}^{\text{q}}_{k,i} = \text{MSFSeg}(\mathbf{I}^{\text{q}}_{k,i},\mathbf{I}^{\text{s}}_{1..n}, \mathbf{M}^{\text{s}}_{1..n})$.
\EndFor
\State Add labeled volume $\hat{V}_k$ to support pool: $\mathcal{P} \gets \hat{V}_k$.
\EndFor
\end{algorithmic}
\end{algorithm}

\section{Experiments}
\label{sec:exp}
We evaluate MSFSeg through various experiments. First, we apply MSFSeg to conventional few-shot segmentation (FSS) with 1-shot setting, in order to compare to prior art, as methods inherently tackling 3D few-shot segmentation are lacking. For 1-shot case, we randomly apply data augmentation techniques, e.g. horizontal flipping, affine transformations or color jittering and input ``n" number of augmented support images (as pseudo n-shot case) to explore comprehensive morphology information from the single support.
For 3D FSS, we highlight the satisfying performance of MSFSeg on our proprietary and challenging dataset of anatomical tubular structures.

\subsection{Evaluation Setup and Protocols}

\noindent\textbf{Implementation Details.}
We apply a ResNet-101 \cite{resnet_CVPR16} pretrained on ImageNet-1K \cite{deng2009imagenet} as feature extractor for fair comparison with prior art, with input image size set to $384\times384$. Please see supplementary material for more architecture and implementation details, experimental results and model parameter analysis.

\noindent\textbf{Datasets.}
To evaluate conventional FSS on medical scans, we follow the same train/test split and evaluation protocol as \cite{ouyang2020self,roy2020squeeze,CRAPNet_WACV23}, and train our network on Abdomen-CT \cite{atlas_dataset} and CHAOS-MRI \cite{CHAOS2021} datasets. Accordingly, we test our proposed MSFSeg in 2 different settings, with \textit{setting 1} using all training/testing data, and \textit{setting 2} enforcing testing classes to be completely unseen by removing any training image that contains a testing class object.
While state-of-the-art FSS methods \cite{ouyang2020self,CRAPNet_WACV23,SRCL_MICCAI22} typically only consider 1-shot setting, we further perform 5-shot evaluations on Abdomen-CT and CHAOS-MRI, randomly selecting 5 support slices across the dataset. 
For few-shot 3D segmentation on our proprietary dataset, we train MSFSeg on domain-irrelevant natural \cite{cocodataset} and medical \cite{CHAOS2021,atlas_dataset} datasets. Our collected test set contains $40$ CT data volumes collected from hospitals, with 4 objects of interest, \ie, \textit{hepatic artery}, \textit{renal artery}, \textit{renal vein}, and \textit{carotid artery}, manually annotated for ground-truth. Note that these objects of interest are all challengingly small tubular structures, unseen during training.

\noindent\textbf{Evaluation Metrics.}
We report Dice scores \cite{ouyang2020self,CRAPNet_WACV23,CAT_ICIT22,SRCL_MICCAI22} on Abdomen-CT/CHAOS-MRI for comparison to prior FSS methods on medical images. We also report Region Jaccard ($\mathcal{J}$), Boundary F measure ($\mathcal{F}$), and their average ($\mathcal{J\&F}$) additional to Dice scores, commonly used as video object segmentation metrics \cite{Pont-Tuset_arXiv_2017,Xu2018YouTubeVOSAL} and are thus suitable for few-shot 3D segmentation evaluation. We report the average over 5 runs for all our experimental results.

\begin{table*}[t]
  \scriptsize
  \caption{Comparison to prior art (Dice score) on various datasets and settings. $\mathcolor{green}{^\Asterisk}$ indicates method using 3D scribble inputs and full 3D labels for network training. Evaluations are performed on left kidney(LK), right kidney(RK), spleen, liver.}
  \centering
{
\resizebox{1.\textwidth}{!}{
\begin{tabular}{c|l| c| c c c c c | c c c c c}
    \toprule
    & \multirow{2}*{Methods} & \multirow{2}*{\makecell{Super-pixel\\or -voxel}} & \multicolumn{5}{c|}{Abdomen-CT \cite{atlas_dataset}} & \multicolumn{5}{c}{CHAOS-MRI \cite{CHAOS2021}} \\
      \cline{3-13}
    & & & LK & RK & Spleen & Liver & avg & LK & RK & Spleen & Liver & avg\\
    \cmidrule{2-13}
    {\multirow{12}{*}{\rotatebox[origin=c]{90}{\textit{\textbf{Setting 1}}}}} & SE-Net{\cite{roy2020squeeze}} & \xmark & 24.42 & 12.51 & 43.66 & 35.42 & 29.00 & 45.78 & 47.96 & 47.30 & 29.02 & 42.51 \\
    & Vanilla PANet{\cite{wang2019panet}} & \xmark & 20.67 & 21.19 & 36.04 & 49.55 & 31.86 & 30.99 & 32.19 & 40.58 & 50.40 & 38.53\\
    & ALPNet{\cite{ouyang2020self}} & \xmark & 29.12 & 31.32 & 41.00 & 65.07 & 41.63 & 44.73 & 48.42 & 49.61 & 62.35 & 51.28\\
    & VQV~\cite{VQV_CVPR23} & \xmark & - & - & - & - & - & 60.03 & 68.94 & {79.08} & {81.72} & 72.44 \\
    \cmidrule{2-13}
    & SSL-PANet$\color{red}$\cite{ouyang2020self} & \cmark & 56.52 & 50.42 & 55.72 & 60.86 & 57.88 & 58.83 & 60.81 & 61.32 & 71.73 & 63.17\\
    & SSL-ALPNet$\color{red}${\cite{ouyang2020self}} & \cmark & 72.36 & 71.81 & 70.96 & {78.29} & 73.35 & 81.92 & 85.18 & 72.18 & 76.10 & 78.84\\
    & CRAPNet{\cite{CRAPNet_WACV23}} & \cmark & {74.69} & {74.18} & 70.37 & 75.41 & 73.66 & {81.95} & {86.42} & {74.32} & {76.46} & 79.79 \\
    & RPT~\cite{RPT_MICCAI23} & \cmark & 77.05 & 79.13 & 72.58 & 82.57 & 77.83 & 80.72 & 89.82 & 76.37 & \textbf{82.86} & 82.44 \\
    & SSL-VQV~\cite{VQV_CVPR23} & \cmark & - & - & - & - & - & \textbf{89.54} & \textbf{91.56} & {77.21} & {79.92} & \textbf{84.56} \\
    \cmidrule{2-13}
    & PRNet$\mathcolor{green}{^\Asterisk}$\cite{scribble_TMI23} {\scriptsize(1 scribble)}& \xmark & - & - & - & - & - & 85.10 & 82.30 & 83.50 & 78.10 & 82.30\\
    \cmidrule{2-13}
    & \textbf{Ours -- 1-shot} (1 slice) & \xmark & \textbf{81.11} & \textbf{78.41} & \textbf{73.64} & \textbf{78.91} & \textbf{78.02} & {84.18} & {88.10} & {77.12} & 76.11 & {81.38} \\
    & \textbf{Ours -- 5-shot} (5 slices) & \xmark & \textbf{87.22} & \textbf{85.62} & \textbf{82.71} & \textbf{82.57} & \textbf{84.53} & {88.63} & {90.94} & \textbf{82.73} & {82.10} & \textbf{86.10} \\
    \midrule
    
    {\multirow{14}{*}{\rotatebox[origin=c]{90}{\textit{\textbf{Setting 2}}}}} & ALPNet-init{\cite{ouyang2020self}} & \xmark & 13.90 & 11.61 & 16.39 & 41.71 & 20.90 & 19.28 & 14.93 & 23.76 & 37.73 & 23.93\\
    & SE-Net{\cite{roy2020squeeze}} & \xmark & 32.83 & 14.34 & 0.23 & 0.27 & 11.91 & 62.11 & 61.32 & 51.80 & 27.43 & 50.66 \\
    & Vanilla PANet{\cite{wang2019panet}} & \xmark & 32.34 & 17.37 & 29.59 & 38.42 & 29.43 & 53.45 & 38.64 & 50.90 & 42.26 & 46.33\\
    & ALPNet{\cite{ouyang2020self}} & \xmark & 34.96 & 30.40 & 27.73 & 43.37 & 35.11 & 53.21 & 58.99 & 52.18 & 37.32 & 50.43\\
    & CAT {\cite{CAT_ICIT22}} & \xmark & - & - & - & - & - & 74.92 & 71.89 & {75.43} & 64.57 & 71.70\\
    & CAT-Net~\cite{CAT_MICCAI23} & \xmark & 63.36 & 60.05 & 67.65 & 75.31 & 66.59  & - & - & - & - & -\\
    \cmidrule{2-13}
    & SSL-PANet{\cite{ouyang2020self}} & \cmark & 37.58 & 34.69 & 43.73 & 61.71 & 44.42 & 47.71 & 47.95 & 58.73 & 64.99 & 54.85\\
    & SSL-ALPNet{\cite{ouyang2020self}} & \cmark & 63.34 & 54.82 & 60.25 & {73.65} & 63.02 & 73.63 & 78.39 & 67.02 & 73.05 & 73.02\\
    & SSL-RPNet{\cite{RPNet_ICCV21}} & \cmark & 65.14 & 66.73 & 64.01 & 72.99 & 67.22 & 71.46 & 81.96 & 73.55 & {75.99} & 75.74 \\
    & CRAPNet{\cite{CRAPNet_WACV23}} & \cmark & {70.91} & {67.33} & {70.17} & 70.45 & {69.72} & 74.66 & 82.77 & 70.82 & 73.82 & 75.52 \\
    & RPT~\cite{RPT_MICCAI23} & \cmark & 72.99 & 67.73 & 70.80 & 75.24 & 71.69 & 78.33 & 86.01 & 75.46 & \textbf{76.37} & 79.04 \\
    & SRCL{\cite{SRCL_MICCAI22}} & \cmark & 67.39 & 63.37 & 67.36 & 73.63 & 67.94 & {77.07} & {84.24} & 73.73 & 75.55 & {77.65} \\
    \cmidrule{2-13}
    & \textbf{Ours -- 1-shot} (1 slice) & \xmark & \textbf{79.24} & \textbf{77.36} & \textbf{75.21} & \textbf{76.73} & \textbf{77.14} & \textbf{82.83} & \textbf{86.98} & \textbf{78.07} & {76.14} & \textbf{81.01} \\
    & \textbf{Ours -- 5-shot} (5 slices) & \xmark & \textbf{85.73} & \textbf{84.51} & \textbf{81.60} & \textbf{81.22} & \textbf{83.27} & \textbf{87.70} & \textbf{90.62} & \textbf{81.97} & \textbf{82.52} & \textbf{85.70} \\
  \bottomrule 
\end{tabular}}
}
\label{tab:fss_sota_med}
\end{table*}


\subsection{Results and Discussion}

\noindent\textbf{Conventional 1-/5-shot Segmentation.}
Tab.~\ref{tab:fss_sota_med} highlights the superiority of our solution compared to the state-of-the-art for 1-shot FSS tasks, in terms of Dice scores, for nearly all 4 organs over both benchmarks and settings. 
Specifically, our proposed 1-shot solution outperforms existing solutions with a reasonable margin under setting 2, \eg, +5.45\% and +1.97\% on Abdomen-CT and CHAOS-MRI dataset respectively over \cite{RPT_MICCAI23} (2nd-best) on average Dice score over 4 organs. 
Note that numerous existing methods (with \cmark) in Tab.\ref{tab:fss_sota_med} relies on an off-the-shelf superpixel or supervoxel model as auxiliary network to pre-group image pixels. Our proposed MSFSeg, \textit{without using any prior knowledge}, still outperforms them on Abdomen-CT (setting 1 and 2) and CHAOS-MRI (setting 2) datasets.
Additionally, Tab.\ ~\ref{tab:fss_sota_med} and Fig.\ \ref{fig:vis_results} presents 5-shot FSS results (5 support slices for each query) on Abdomen-CT and CHAOS-MRI, highlighting the ability of our proposed MSF to learn comprehensive query-support morphology correlations across multiple supports (\eg, +6.51/+6.13\% average Dice score improvement for 5-shot over 1-shot on Abdomen-CT, for setting 1/2 respectively).
Specifically, compared to the few-shot 3D segmentation method PRNet \cite{scribble_TMI23}, which requires 3D scribbles on data volumes, full 3D labels as data pairs for training, and a heavier architecture; MSFSeg performs significantly better with 5 support 2D slices and comparably with 1 support slice on CHAOS-MRI.


\begin{table*}[t]
  \scriptsize
  \caption{Evaluation of intra-/inter-volume 3D segmentation on challenging tubular targets in proprietary CT data, compared to baseline \cite{hong2022cost}.}
  \centering
  \resizebox{\textwidth}{!}{
    \begin{NiceTabular}{@{\hspace{.0pc}}c|C{1.4cm}|>{\hspace{.07pc}}c>{\hspace{.07pc}}c<{\hspace{.07pc}}c<{\hspace{.07pc}}c<{\hspace{.07pc}}|>{\hspace{.07pc}}c>{\hspace{.07pc}}c<{\hspace{.07pc}}c<{\hspace{.07pc}}c<{\hspace{.07pc}}|>{\hspace{.07pc}}c>{\hspace{.07pc}}c<{\hspace{.07pc}}c<{\hspace{.07pc}}c<{\hspace{.07pc}}|>{\hspace{.07pc}}c>{\hspace{.07pc}}c<{\hspace{.07pc}}c<{\hspace{.07pc}}c<{\hspace{.07pc}}}
    \toprule
    \diagbox{}{}  &   
    Metrics$\rightarrow$ & 
    \multicolumn{4}{c|}{$\mathcal{J}$} & \multicolumn{4}{c|}{$\mathcal{F}$} & \multicolumn{4}{c|}{$\mathcal{J\&F}$} & \multicolumn{4}{c}{Dice} \\
\cmidrule{1-18}   
    Target$\downarrow$ & 
\diagbox{\scriptsize propag.}{\scriptsize $n$-shot} 
    & {\scriptsize baseline}  & $n$=1 & $n$=5 & $n$=$5^2$ & base.  & $n$=1 & $n$=5 & $n$=$5^2$ & base. & $n$=1 & $n$=5 & $n$=$5^2$ & base.  & $n$=1 & $n$=5 & $n$=$5^2$ \\
    \midrule
    \multirow{2}[1]{*}{Hep. A.} & 
    intra-vol. & .293 & .559 & .629 & .640 & .394 & .656 & .672 & .678 & .344 & .607 & .651 & .659 & .393 & .709 & .757 & .767
    \\
    & inter-vol. & .283 & .580 & .625 & .635 & .387 & .634 & .662 & .667 & .335 & .606 & .643 & .651 & .383 & .688 & .753 & .760
    \\
\cmidrule{1-18} 
    \multirow{2}[1]{*}{Ren. A.} & 
    intra-vol. & .155 & .610 & .608 & .647 & .266 & .745 & .814 & .830 & .210 & .677 & .711 & .738 & .212 & .678 & .726 & .763
    \\
    & inter-vol. & .137 & .510 & .580 & .638 & .248 & .780 & .798 & .826 & .192 & .645 & .689 & .732 & .193 & .635 & .696 & .753
    \\
\cmidrule{1-18} 
    \multirow{2}[1]{*}{Ren. V.} & 
    intra-vol. & 
    .137 & .395 & .487 & .519 & .287 & .569 & .700 & .706 & .212 & .482 & .593 & .612 &.213 & .508 & .624 & .655
    \\
    & inter-vol. & 
     .128 & .348 & .502 & .491 & .277 & .595 & .686 & .697 & .203 & .471 & .594 & .594 & .203 & .474 & .608 & .627
    \\
\cmidrule{1-18} 
    \multirow{2}[1]{*}{Car. A.} & 
    intra-vol. & .233 & .393 & .430 & .479 &  .313 & .554 & .570 & .609 & .273 & .473 & .500 & .544 & .295 & .491 & .536 & .583
    \\
    & inter-vol. & 
    .232  & .389 & .393 & .459 & .312  & .516 & .552 & .599 & .272  & .452 & .472 & .529 & .294  & .447 & .501 & .569 \\
    \bottomrule
    \end{NiceTabular}}
\label{tab:3D_3D_intra}
\end{table*}


\begin{table}[t]
\parbox{.55\linewidth}{
    \centering
    \includegraphics[width=\linewidth]{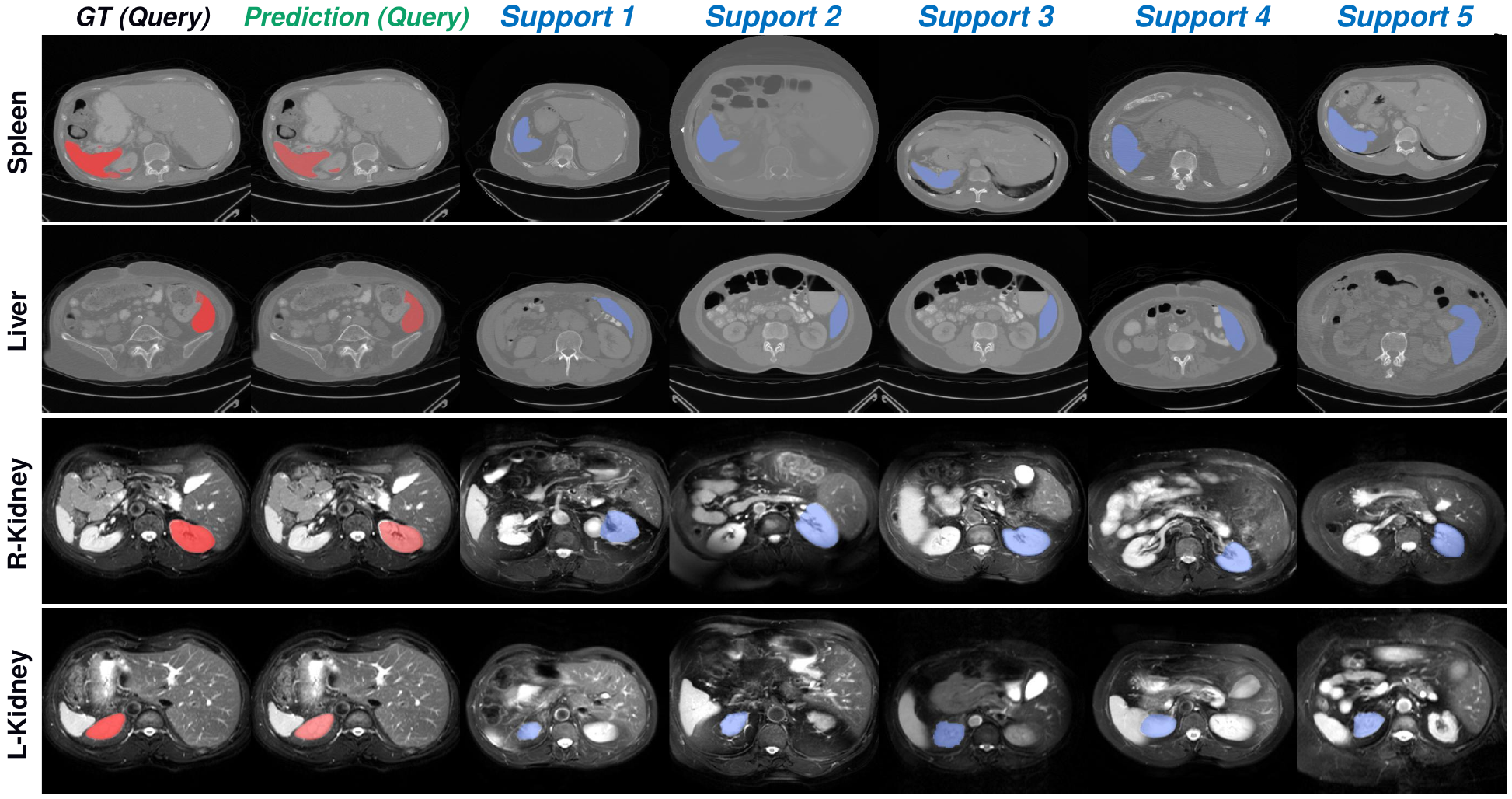}
    \captionof{figure}{
    Qualitative results of 5-shot FSS. 
    }
    \label{fig:vis_results}
}
\hfill
\parbox{.43\linewidth}{
    \vspace{-2em}
    \centering
    \caption{MSF ablation study on CHAOS-MRI (5-shot, setting 2).}
    \resizebox{.43\textwidth}{!}{
    \begin{tabular}{cccc|c | c | c | c}
    \toprule
    $\mathbf{\hat{M}}^{\text{q}}_{\text{coh}}$ & $\mathbf{\hat{M}}^{\text{q}}_{\text{div}}$ & $\mathbf{\hat{M}}^{\text{q}}_{\text{ca}}$ & $\mathbf{\hat{M}}^{\text{q}}_{\text{avg}}$ & LK & RK & Spleen & Liver \\ 
    \midrule
    \cmark & \xmark & \xmark & \xmark & 84.81 & 88.64 & 73.94 & 81.47 \\ 
    \xmark & \cmark & \xmark & \xmark & 81.11 & 87.16 & 71.58 & 82.14 \\
    \cmark & \cmark & \xmark & \xmark & 85.74 & 90.20 & 78.75 & 82.33 \\
    \midrule
    \xmark & \xmark & \cmark & \xmark & 86.28 & 89.46 & 81.14 & 70.66 \\
    \xmark & \xmark & \xmark & \cmark & 87.12 & 89.78 & 79.85 & 79.33 \\
    \xmark & \xmark & \cmark & \cmark & 87.17 & 90.11 & 81.30 & 81.90 \\
    \midrule
    \cmark & \cmark & \cmark & \cmark & \textbf{87.70} & \textbf{90.62} & \textbf{81.97} & \textbf{82.52} \\
    \bottomrule
    \end{tabular}
    }
    \label{tab:ablation}%
}
\end{table}

    

\noindent\textbf{Few-Shot 3D Segmentation.}
We evaluate our few-shot 3D segmentation workflow on both intra-volume and inter-volume cases and report the scores over all data volumes for each object of interest in Tab.~\ref{tab:3D_3D_intra}. 
To contrast MSFSeg's superiority, we apply a state-of-the-art FSS method, VAT \cite{hong2022cost} (retrained/evaluated with same training/testing splits as our protocol), as baseline.
For 1-shot intra-volume inference, we consider the central slice and corresponding ground-truth label as support and sequentially infer the other masks in the volume, whereas for 5-shot segmentation, we randomly select 5 support slices for each query slice. 
We further demonstrate MSFSeg ability to process 3D/sequential supports by randomly selecting 5 sequences as support, each sequence containing 5 consecutive slices (``$n$=$5^2$'' in Tab.~\ref{tab:3D_3D_intra}).
The inter-volume segmentation is performed as described in Alg.\ \ref{algo:prop_workflow}, using 1 or 5 most similar sequences in the labeled pool. 
Despite its similar backbone, VAT cannot properly propagate the fine labels of new tubular objects, whereas our method yields satisfying results \wrt both intra- and inter-volume 3D segmentation thanks to its fusion module and cross-volume workflow. 
Tab.\  \ref{tab:3D_3D_intra} also shows how MSF leverages additional support data for improved accuracy ($\geq$6\% mean Dice score increase from $n=1$ to 5). The small margin performance gap between intra- and inter-volume segmentation also proves the generalizability of MSFSeg; \ie, support across sequences from different patient scans does not compromise segmentation results compared to label support from ground truth (user).

\noindent\textbf{Ablation Study of Multi-Surrogate Fusion.}
We demonstrate the role of each MSF surrogate (coherence, diverse, attention and stabilization operations) through an ablation study on CHAOS-MRI in Tab.\ \ref{tab:ablation} (5-shot, setting 2). The study confirms the positive impact of each surrogate function and the combination of local (coherence/diverse) and global (attention/stabilization) expertise.


\section{Conclusion}
We proposed MSFSeg, a novel solution for segmenting unseen (cross-domain) 3D objects given only a few labeled slices.
Our solution rely on a lightweight 2D low-shot segmentation network with multi-surrogate fusion to aggregate information from multiple supports for 3D segmentation in a transferred domain, and on a pool-based workflow to select the most relevant 2D/3D candidates for cross-volume generalization. 
MSFSeg can be integrated into actual 3D medical annotation tools to greatly reduce the burden of medical data annotators by proposing quick and valid initial segmentation masks across whole patient populations, as demonstrated on unseen datasets with challenging anatomical targets.

\begin{credits}
\subsubsection{\discintname}
The authors have no competing interests to declare that are
relevant to the content of this article. 
\end{credits}

\bibliographystyle{splncs04}
\bibliography{egbib}

\end{document}